\documentclass[10pt,twocolumn,letterpaper]{article}

\usepackage{iccv}
\usepackage{times}
\usepackage{epsfig}
\usepackage{graphicx}
\usepackage{amsmath}
\usepackage{amssymb}
\usepackage{xcolor}
\usepackage{bbm}
\usepackage{algorithm} 
\usepackage{algpseudocode}
\usepackage{multirow}
\usepackage{makecell} 
\usepackage{authblk}
\usepackage{booktabs}
\usepackage{dcolumn}
\algrenewcommand\algorithmicrequire{\textbf{Input:}}
\algrenewcommand\algorithmicensure{\textbf{Output:}}

\usepackage{hyperref}
\hypersetup{pagebackref=true,breaklinks=true,letterpaper=true,colorlinks,bookmarks=false}

\iccvfinalcopy 

\def\iccvPaperID{} 

\ificcvfinal\fi

\usepackage{xcolor}

\begin{document}

\title{Development of a deep learning-based tool to assist wound classification}

\author[1]{Po-Hsuan Huang, M.S.}
\author[2]{Yi-Hsiang Pan, M.D.}
\author[1]{Ying-Sheng Luo, M.S.}
\author[3,4]{Yi-Fan Chen, M.D.}
\author[5,6]{Yu-Cheng Lo, M.D.}
\author[1,7]{Trista Pei-Chun Chen, Ph.D. \thanks{Work done while the author was affiliated with Inventec Corporation}}
\author[2,8,9]{Cherng-Kang Perng, M.D., Ph.D.\thanks{Corresponding author. E-mail address: ckperng@vghtpe.gov.tw}}

\affil[1]{\small Inventec AI Center, Inventec Corporation, Taipei, Taiwan}
\affil[2]{\small Division of Plastic and Reconstructive Surgery, Department of Surgery, Taipei Veterans General Hospital, Taipei, Taiwan}
\affil[3]{\small Division of Plastic and Reconstructive Surgery, Department of Surgery, Taipei Medical University Hospital}
\affil[4]{\small Department of Surgery, School of Medicine, College of Medicine, Taipei Medical University, Taipei, Taiwan}
\affil[5]{\small Center for Quality Management, Taipei Veterans General Hospital, Taipei, Taiwan}
\affil[6]{\small Institute of Bio-Medical Informatics, National Yang-Ming Chiao Tung University, Taipei, Taiwan}
\affil[7]{\small AI Research Center, Microsoft Corporation, Taipei, Taiwan}
\affil[8]{\small Department of Surgery, School of Medicine, National Yang Ming Chiao Tung University, Taipei, Taiwan}
\affil[9]{\small Chang Bing Show Chwan Memorial Hospital, Changhua, Taiwan}

\maketitle
\ificcvfinal\thispagestyle{empty}\fi

\begin{abstract}
This paper presents a deep learning-based wound classification tool that can assist medical personnel in non-wound care specialization to classify five key wound conditions, namely deep wound, infected wound, arterial wound, venous wound, and pressure wound, given color images captured using readily available cameras. The accuracy of the classification is vital for appropriate wound management. The proposed wound classification method adopts a multi-task deep learning framework that leverages the relationships among the five key wound conditions for a unified wound classification architecture. With differences in Cohen’s kappa coefficients as the metrics to compare our proposed model with humans, the performance of our model was better or non-inferior to those of all human medical personnel. Our convolutional neural network-based model is the first to classify five tasks of deep, infected, arterial, venous, and pressure wounds simultaneously with good accuracy.  The proposed model is compact and matches or exceeds the performance of human doctors and nurses. Medical personnel who do not specialize in wound care can potentially benefit from an app equipped with the proposed deep learning model.

\end{abstract}
\section{Introduction}\label{sec:intro}

Wound healing is one of the oldest and yet the most important challenges in the medical field. Normal wound healing is a dynamic, interactive process involving soluble mediators, blood cells, extracellular matrix, and parenchymal cells~\cite{ref1_singer1999cutaneous}. Factors that deviate from the normal process will result in abnormal healing and chronic wounds. Chronic wounds affect the quality of life of nearly 2.5\% of the total population in the United States~\cite{ref2_sen2021human}. The management of wounds has a significant economic impact on health care; this becomes even more challenging with aging of the population. The underlying mechanism of chronic wounds varies significantly, but includes factors that influence blood supply (peripheral vascular disease), immune function (such as immunosuppression or acquired immunodeficiency), metabolic diseases (such as diabetes), medications, or previous local tissue injury (such as radiation therapy)~\cite{ref3_han2017chronic}. Well-trained medical wound specialists are crucial for correct diagnosis and proper wound treatment, but are usually not readily available in primary healthcare facilities.

The increasing use of artificial intelligence (AI) technologies and portable devices such as smartphones has led to timely development of remote and intelligent diagnosis and prognosis systems for wound care~\cite{ref4_anisuzzaman2022image}. Deep learning, a subdomain of machine learning and AI, was inspired by the human brain and requires a large amount of data for automatic mapping between the input and output. Certain pre-set rules are not required. Image-based wound classification using deep learning is a new field of interest. Convolutional neural network (CNN) -based methods have been proposed to detect infected wounds~\cite{ref5_wang2015unified,ref6_shenoy2018deepwound}, identify the feature differences between a healthy skin and diabetic foot wounds~\cite{ref7_goyal2018dfunet}, and perform optimized segmentation of different tissue types present in pressure injuries~\cite{ref8_zahia2018tissue}.

Well-trained medical wound specialists are crucial for correct diagnosis and proper wound treatment. However, they are usually not readily available in primary healthcare facilities. An intelligent system can provide information about patients’ wounds to general practitioners, nurses, and even the patients themselves before seeing a wound specialist such that timely and efficient referrals can be achieved. We chose five wound assessment tasks to be classified using our deep learning model, namely deep wound, infected wound, arterial wound, venous wound, and pressure wound. Deep wounds are closely related to wound severity. Infected wounds are the most common cause of poor wound healing. Arterial, venous, and pressure wounds are among the most important pathogenesis of chronic non-healing wounds.
\section{Proposed Method}\label{sec:method}
\subsection{Patients and datasets}\label{ssec:patients_datasets}
\begin{figure}[t!]
\centerline{
	\hspace{3mm}\includegraphics[width=1.0\columnwidth]{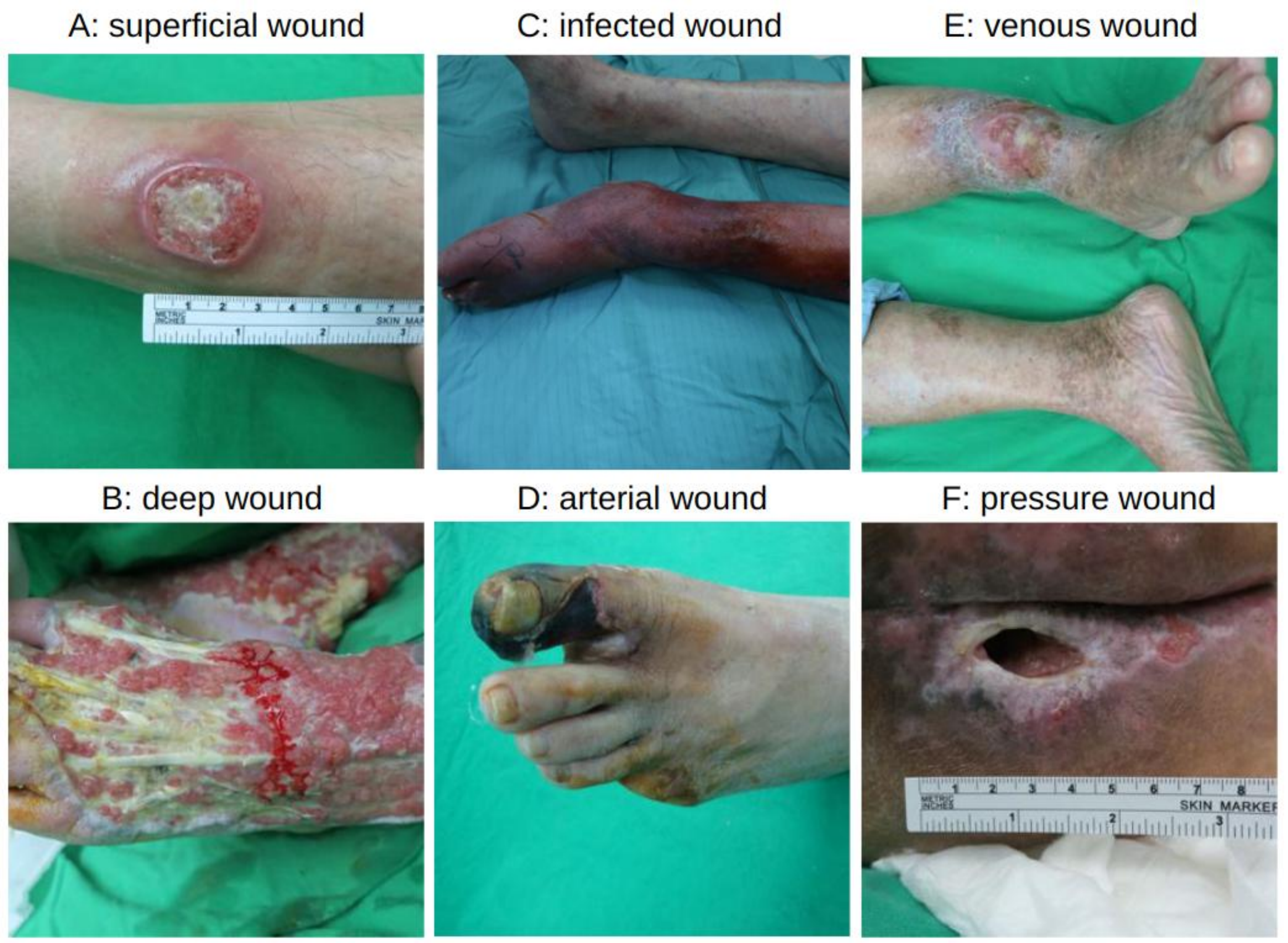}
}  
    \caption{Some sample images from our dataset. A: superficial wound, B: deep wound, C: infected wound, D: arterial wound, E: venous wound, F: pressure wound.}
    \vspace{-2mm}
	\label{fig:Figure1}
\end{figure}

Our dataset consists of 2149 wound images collected from 1429 patients randomly selected from the database of the corresponding author’s affiliated hospital. The 1429 patients consist of 900 men and 529 women with a mean age of 62.6 years and a standard deviation of 18.5 years. This study was approved by the Institutional Review Board (approval number: 2020-08-008A). The wound images in our dataset were taken by medical personnel under various imaging conditions such as lighting, capturing devices, shooting angles, resolutions, and backgrounds (some with irrelevant surrounding objects). Five classification tasks were included (Figure~\ref{fig:Figure1}): deep wound, infected wound, arterial wound, venous wound, and pressure wound. All the wound images were reviewed by a medical wound specialist. Each task is a binary classification with either a negative or a positive label. A negative label in deep wound indicates no wound or shallower than deep fascia; a positive label in deep wound indicates that the wound is deeper than deep fascia. For the other four tasks, a negative label indicates the absence of such symptoms. In designing the deep learning model as well as in the experiments, we partitioned the dataset by patients, and not images, because a patient may have more than one wound image in the dataset. That is, some patients were used as training subjects, whereas the rest were used for testing. The dataset was randomly shuffled and divided into training (80\% of all patients) and testing (20\% of all patients) datasets. Table~\ref{table:datasets} presents the data distributions of all the images and for each task.

\subsection{Model architecture and implementation details}\label{ssec:model architectur}
A conventional machine learning-based method requires two steps to classify an image. The first step is to find the relevant segment from the image or perform feature extraction. Next, a classifier, such as a support vector machine, performs classification on the extracted image segment or features. Alternatively, deep learning-based approaches learn features and construct classifiers within their deep learning model without explicitly separating the steps. Our proposed method belongs to the latter category. 

In this study, we build a DL-based CNN model to perform five wound classification tasks. A CNN-based classification model performs steps of feature extraction from the original input image to gradually extract lower-level features such as shapes and texture to higher-level features such as semantic information, all automatically learned from the training data. The extracted features are then used to classify the input image.

\begin{figure}[t!]
\centerline{
	\hspace{3mm}\includegraphics[width=1.0\columnwidth]{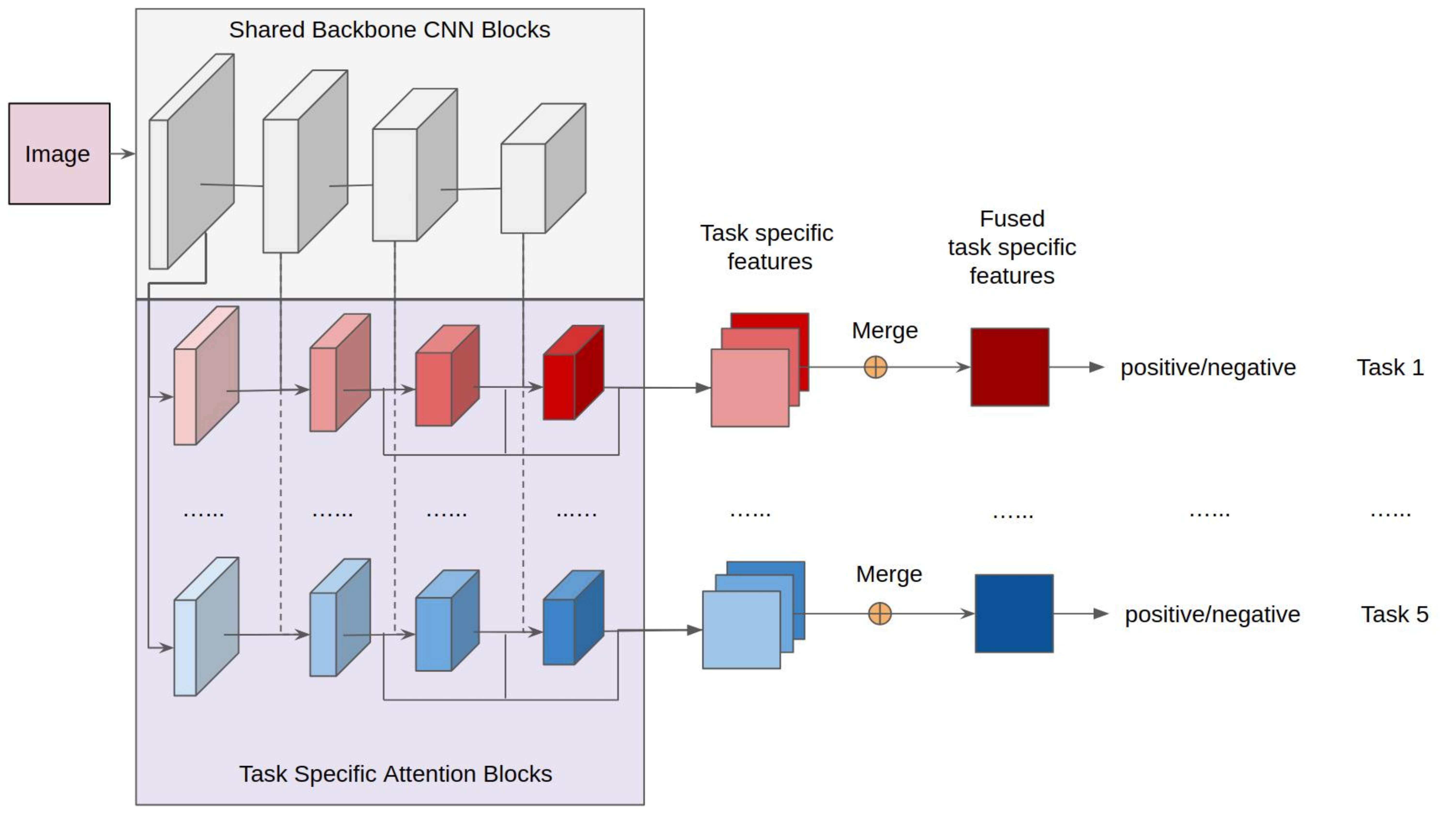}
}  
    \caption{Overview of our wound assessment model framework. The grey blocks include the shared backbone blocks. The purple regions contain task-specific attention blocks. Each task has one branch; the model has a total of five individual task branches. We only showed two branches here for clarity. Each task branch uses task-specific attention blocks to extract shared features from shared backbone blocks. The task branch outputs feature maps from different levels and merge them into fused task-specific features. The classifier then accepts the features and outputs the prediction.}
    \vspace{-2mm}
	\label{fig:Figure2}
\end{figure}

It is possible to design five separate deep learning models to tackle the task separately. However, this could be expensive in terms of computation and memory requirements. We observed that these tasks may share meaningful features, such as skin texture or localization information of the wound. To leverage the intrinsic relationships among them, we designed a CNN-based model that learns these five tasks simultaneously and also provides the results simultaneously. There are many different ways to share features among these tasks~\cite{ref9_ruder2017overview,ref10_misra2016cross,ref11_liu2019end,ref12_gao2019nddr,ref13_meyerson2017beyond,ref14_sener2018multi,ref15_zhang2021survey,ref16_he2016deep,ref17_deng2009imagenet}. Figure~\ref{fig:Figure2} shows an overview of our wound classification model. Following~\cite{ref11_liu2019end}, all five tasks in our model shared the same backbone model with their task-specific branches. Those task-specific branches are composed of attention modules that imitate the human attention mechanism. They help each task branch extract its critical features. In addition, we aggregate different levels of information to improve the final classification performance.



\begin{table}[t]
\caption{Data distribution of our 1429 patients (2149 images). We show the number of positive images of each task in the training, validation, and test datasets. The percentage of positive images of each task is in a bracket under the total number of images of the corresponding task.}
\label{table:datasets}
\centering
\resizebox{\columnwidth}{!}{
\begin{tabular}{@{}cccccc@{}}
\toprule
\multicolumn{2}{l}{}
& \multicolumn{3}{c}{Dataset}
& \\
\cmidrule(lr){3-5}
\multicolumn{2}{c}{}
& Training
& Validation
& Test
& Total \\
\midrule
\multirow{2}{*}{{\makecell{Number \\in dataset}}}
& Patient & 979 & 164 & 286 & 1429 \\
& Image & 1498 & 225 & 426 & 2149 \\
\midrule
\multirow{5}{*}{\makecell{Number of \\positive images \\(percentage)}}
& Deep wound & 1006 & 134 & 251 & \makecell{1391 \\(64.7 \%)} \\
& Infected wound & 903 & 126 & 259 & \makecell{1288 \\(59.9 \%)} \\
& Arterial wound & 316 & 42 & 96 & \makecell{454 \\(21.1 \%)} \\
& Venous wound & 26 & 7 & 19 & \makecell{52 \\(2.4 \%)} \\
& Pressure wound & 171 & 36 & 60 & \makecell{267 \\(12.4 \%)} \\
\bottomrule
\end{tabular}
}
\end{table}

To address the class imbalance problem in our dataset, we used the weighted cross-entropy loss as the loss function to optimize our model. The smaller the data size of a class, the larger the class weight, so that the model would not be biased to a class because of its data size. To increase the robustness of the model and reduce overfitting to certain imaging conditions, we applied training image data augmentation by performing realistic affine transformations, vertical or horizontal flipping, brightness altering, and contrast adjustments. This augmentation strategy is to mimic the different imaging conditions which are beneficial to model learning. The augmentation is carefully chosen and will not produce unrealistic data. For example, real images are sometimes upside-down or taken from a different angle. A human can rotate the image manually or can identify the image regardless of the upside-down. To imitate this ability, we apply vertical flipping to the original image and use both versions of images to train the model during the training process. The model will learn from both images to become more robust and knows that the upside-down is irrelevant to the disease. Note that we do not perform augmentations on testing data. Given the varying original image sizes, we manually cropped them to a square shape and resized them to the minimal size of all cropped images, which is 300 × 300 pixels. 

\subsection{Model evaluation and statistical analysis}\label{ssec:Model evaluation}

To evaluate the effectiveness of our proposed deep learning model, we used some performance evaluation metrics from descriptive statistics, including accuracy, sensitivity, specificity, and area under the receiver operating characteristic curve (AUC). AUC is particularly effective for assessing classifiers trained with highly imbalanced data. Given the nature of our data, we included this metric in addition to the rest of the performance evaluation metrics. 

Furthermore, it is worthwhile to compare the clinical performance of the proposed deep learning model with that of medical personnel.  Seven human subjects participated in this study, including two surgical attendings specialized in wound care, one senior surgical resident, one junior resident, one nurse practitioner with wound specialty, and two registered nurses of the surgical ward. In the experiments, we labeled them as attending A, attending B, resident A, resident B, nurse A, nurse B, and nurse C, respectively. We randomly selected 350 images from the test dataset with a similar distribution as the test set for all the experiments, including the deep learning model and seven human medical personnel. This is referred to as a smaller test dataset “human test dataset”. The medical personnel were provided with images in the human test dataset and were asked to answer if each image is positive for all five wound conditions. To compare the performance of the proposed deep learning model with those of human subjects, we used the accuracy, sensitivity, specificity, and Cohen’s kappa coefficient~\cite{ref18_mchugh2012interrater}. Cohen’s kappa coefficient is commonly used to measure the rater reliability for categorical items. Cohen’s kappa coefficient was calculated with respect to the ground truth labeled by the medical wound specialist. The higher the Cohen’s kappa coefficient, the better the performance.

We used the difference in the Cohen’s kappa coefficient values (the proposed deep learning model minus medical personnel) with their 95\% confidence interval (CI) as the metric for the comparison. The 95\% CI is calculated using bootstrapping~\cite{ref19_efron1994introduction}. The proposed model is significantly better if such a difference is positive and its CI does not cross zero. It is non-inferior if the CI crosses zero, and inferior otherwise. 
We plotted the receiver operating characteristic (ROC) curve~\cite{ref20_mcclish1989analyzing} of our model and calculated the 95\% confidence band around the ROC curve using vertical averaging~\cite{ref21_provost1998case}. The characteristic points of sensitivity and specificity for all medical personnel were calculated and plotted together with the model ROC curve in the same figure for comparison. A side-by-side comparison of the characteristic points and the ROC curve provides another illustrative method for comparison.

\section{Experiments}

\subsection{Performance of the proposed model}
The accuracy, sensitivity, specificity, and AUC of our model using the test dataset are presented in Table~\ref{table:metrics}. In the deep wound task, our model achieved an accuracy of 0.739, sensitivity of 0.693, specificity of 0.806, and AUC of 0.804. In terms of infected wound, the corresponding values for our model are 0.685, 0.653, 0.737, and 0.751, respectively. For arterial wound, the corresponding values for our model are 0.864, 0.688, 0.915, and 0.897, respectively. For venous wound, the corresponding values for our model are 0.960, 0.632, 0.975, and 0.924, respectively. In terms of pressure wound, the corresponding values for our model are 0.906, 0.750, 0.932, and 0.940, respectively.
\begin{table}[t]
\caption{Accuracy, sensitivity, specificity, and AUC of our model for the test dataset.}
\label{table:metrics}
\centering
\vspace{0.2mm}
\resizebox{\columnwidth}{!}{
\begin{tabular}{@{}ccccc@{}}
\toprule
 & Accuracy & Sensitivity & Specificity & AUC \\
\midrule
Deep wound & 0.739 & 0.693 & 0.806 & 0.804 \\
Infected wound & 0.685 & 0.653 & 0.737 & 0.751 \\
Arterial wound & 0.864 & 0.688 & 0.915 & 0.897 \\
Venous wound & 0.960 & 0.632 & 0.975 & 0.924 \\
Pressure wound & 0.906 & 0.750 & 0.932 & 0.940 \\
\bottomrule
\end{tabular}
}
\vspace{-2mm}
\end{table}

\subsection{Performance comparison with the humans}
\begin{figure*}[t!]
\centerline{
	\hspace{3mm}\includegraphics[width=1.0\textwidth]{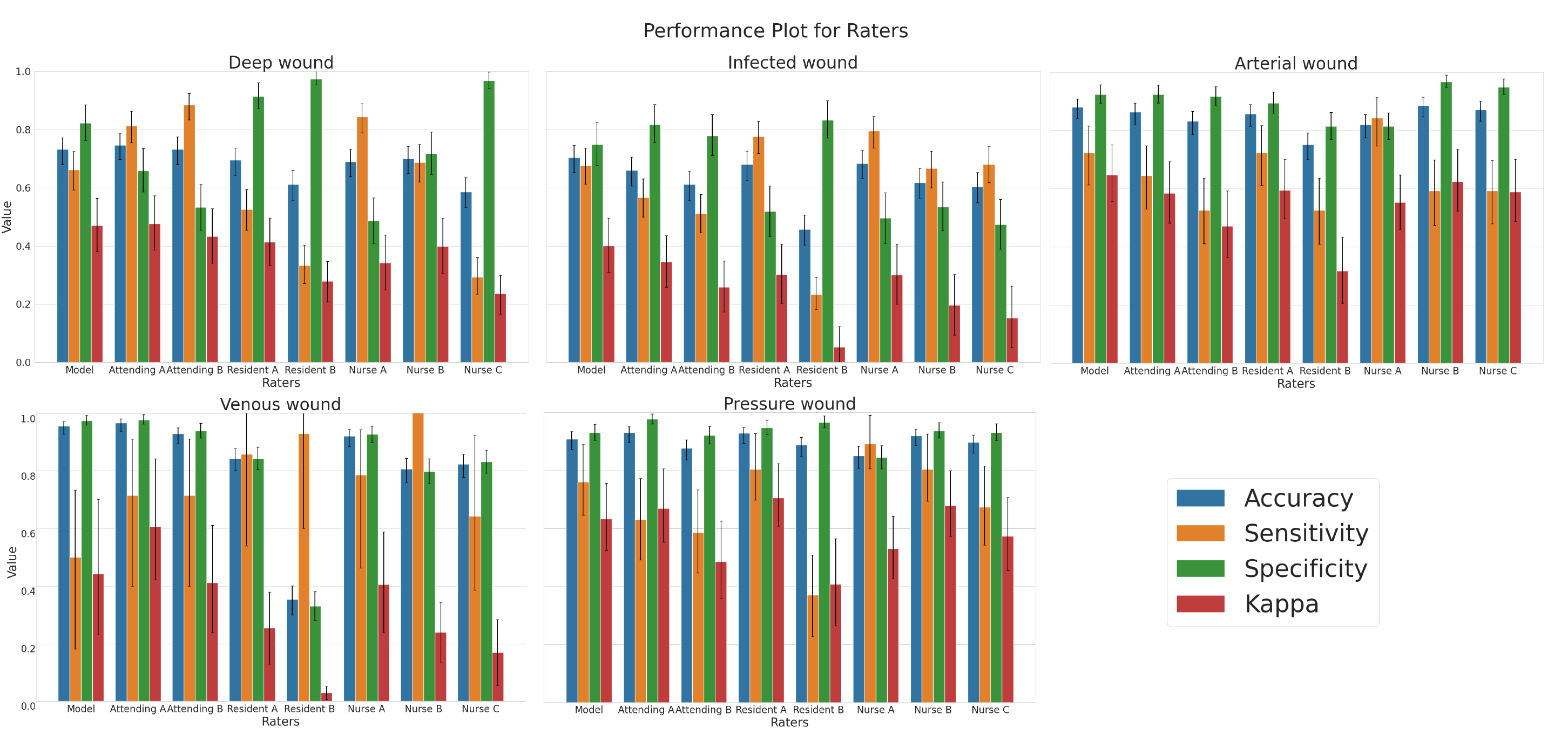}
}  
    \caption{The accuracy, sensitivity, specificity, and Cohen’s kappa with 95\% CI for the proposed model and seven medical personnel.}
    \vspace{-2mm}
	\label{fig:Figure3}
\end{figure*}

The accuracy, sensitivity, specificity, and Cohen’s kappa with 95\% CI for the proposed model and seven medical personnel are showed in Figure~\ref{fig:Figure3}. For the deep wound task, resident A, resident B, and nurse C exhibited low sensitivity values of 0.525, 0.333, and 0.293, respectively. Nurse A exhibited a low specificity of 0.487. For infected wound, our model and all medical personnel yielded the worst average performance. Attending B and resident B exhibited a low sensitivity of 0.511, and 0.233, respectively. Resident A, nurse A, nurse B, and nurse C exhibited low specificity values of 0.517, 0.495, 0.533, and 0.472, respectively.  For arterial wound, attending B, resident B, nurse B, and nurse C exhibited low sensitivity values of 0.526, 0.526, 0.591, and 0.591, respectively. For venous wound, our model yielded a low sensitivity of 0.504. Resident B exhibited a low specificity of 0.330. In terms of pressure wound, resident B exhibited a low sensitivity of 0.370.

It can be observed that the proposed model performed significantly better than resident B, nurse A, and nurse C in the task of deep wound and is non-inferior to the other four medical personnel based on the Cohen’s kappa difference listed in Table~\ref{table:kappa}. In the task of infected wound, our model performed significantly better than attending B, resident B, nurse B, and nurse C and is non-inferior to the others. In the task of arterial wound, our model performed significantly better than attending B, and resident B and is non-inferior to the others. For venous wound, our model performed significantly better than resident B and nurse C and is non-inferior to the others. For pressure wound, our model performed significantly better than resident B and is non-inferior to the others.  

Figure~\ref{fig:Figure4} shows the ROC curves with their 95\% confidence band and the characteristic points of the medical personnel. In the task of deep wound, all the points fell inside the confidence band, implying comparable performance. In the task of infected wound, the characteristic points of attending A, attending B, resident A, and nurse A fell inside the confidence band (implying comparable performance), whereas the rest of the points are on the lower right-hand side of the ROC curve (implying inferior performance to the model). In the tasks of arterial wound and pressure wound, the characteristic points of attending B and resident B fell on the lower right-hand side of the ROC confidence band, implying inferior performance to the model, whereas the characteristic points of the rest of the medical personnel fell inside the ROC confidence band. In the task of venous wound, the characteristic point of the resident B is on the lower right-hand side of the ROC curve (implying inferior performance to the model). From the relative positions of the characteristic points and ROC confidence band, our proposed model is either superior or has a comparable performance as human medical personnel.
\begin{table}[t]
\caption{Difference in Cohen’s kappa with 95\% confidence intervals for performance comparison of our model and medical personnel. * indicates significant difference.}
\label{table:kappa}
\centering
\resizebox{\columnwidth}{!}{
\begin{tabular}{@{}ccccccc@{}}
\toprule

& & Deep wound & Infected wound & Arterial wound & Venous wound & Pressure wound \\
\midrule
\multirow{2}{*}{Attending A} & Difference & -0.007 & 0.057 & 0.062 & -0.162 & -0.037 \\
&95\% CI & [-0.117,0.107] & [-0.045,0.156] & [-0.075,0.194] & [-0.459,0.130] & [-0.192,0.125]\\
\multirow{2}{*}{Attending B} & Difference & 0.035 & \textbf{0.142*} & \textbf{0.177*} & 0.032 & 0.148 \\ 
&95\% CI & [-0.070,0.141] & [0.033,0.253] & [0.045,0.313] & [-0.209,0.266] & [-0.008,0.307] \\
\multirow{2}{*}{Resident A} & Difference & 0.053 & 0.101 & 0.051 & 0.186 & -0.073 \\
&95\% CI & [-0.053,0.159] & [-0.026,0.225] & [-0.069,0.167] & [-0.044,0.403] & [-0.191,0.052]\\
\multirow{2}{*}{Resident B} & Difference & \textbf{0.189*} & \textbf{0.347*} & \textbf{0.328*} & \textbf{0.408*} & \textbf{0.223*} \\
&95\% CI & [0.085,0.291] & [0.242,0.446] & [0.199,0.456] & [0.162,0.632] & [0.036,0.406]\\
\multirow{2}{*}{Nurse A} & Difference & \textbf{0.126*} & 0.101 & 0.093 & 0.037 & 0.101 \\
&95\% CI & [0.012,0.247] & [-0.018,0.222] & [-0.008,0.189] & [-0.216,0.296] & [-0.018,0.219] \\
\multirow{2}{*}{Nurse B} & Difference & 0.07 & \textbf{0.204*} & 0.024 & 0.200 & -0.048 \\ 
&95\% CI & [-0.047,0.186] & [0.080,0.332] & [-0.083,0.132] & [-0.037,0.431] & [-0.190,0.087] \\
\multirow{2}{*}{Nurse C} & Difference & \textbf{0.232*} & \textbf{0.249*} & 0.060 & \textbf{0.269*} & 0.06 \\
&95\% CI & [0.139,0.322] & [0.120,0.376] & [-0.045,0.165] & [0.073,0.456] & [-0.096,0.221]\\

\bottomrule
\end{tabular}
}
\end{table}


\section{Discussion and Conclusions}

The retrospectively collected dataset in our study consisted of wound images taken by medical personnel for the purpose of medical documentation. Such documentation does not impose strict requirements on imaging conditions such as lighting and image capturing devices. Although not as perfect as prospectively collected images with standardized conditions, these images were much closer to images captured in real-world scenarios. The deep learning model derived from these real-world images showed a significantly better or non-inferior performance compared with other medical personnel on all five tasks. Our model should be suitable for further application in primary medical facilities with no strict limitations on imaging devices and conditions.

\begin{figure*}[t!]
\centerline{
	\hspace{3mm}\includegraphics[width=0.9\textwidth]{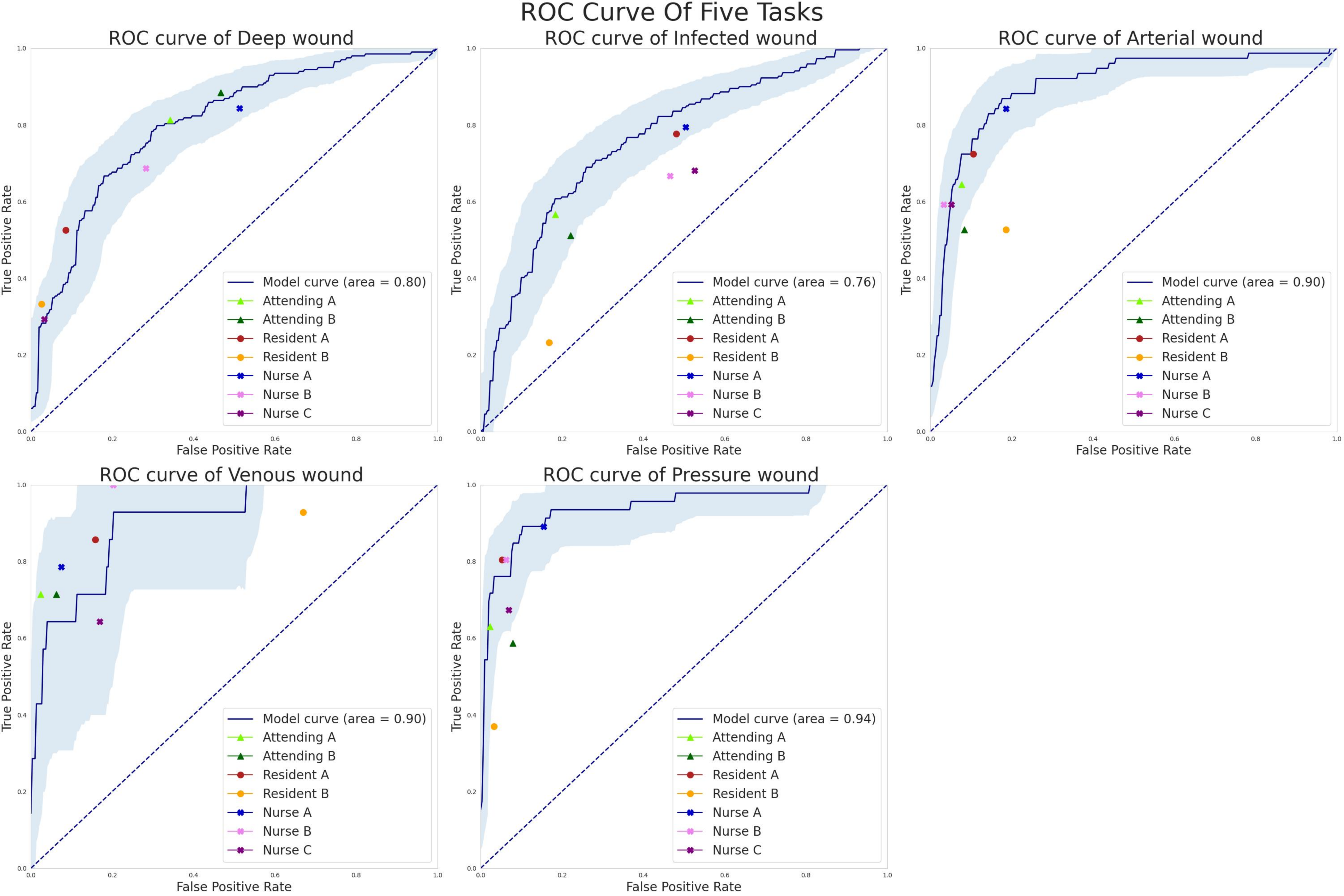}
}  
    \caption{ROC curve of our model, and sensitivity and specificity characteristic points of medical personnel.}
    \vspace{-2mm}
	\label{fig:Figure4}
\end{figure*}

The proposed deep learning model performed well in classifying arterial, venous, and pressure wounds. It exhibited an accuracy and AUC of well above 85\%. The performance is not as profound in the tasks of classifying deep wound and infected wound, with an accuracy of less than 75\% and AUC between 75\% and 80\% for both tasks. It should be noted that in the task of venous wound, our model exhibited a low sensitivity value of 63.2\% and yielded a wide ROC confidence band, as shown in Figure~\ref{fig:Figure4}. This is due to the highly imbalanced nature of our venous wound dataset. There were only 52 positive cases of venous wound in our entire dataset. Aside from the low sensitivity in venous ulcer, we can also observe that the accuracy, specificity, and AUC of the model is usually higher than the sensitivity in all five tasks. This is due to the relatively smaller number of positive cases compared to negative cases in the training data for all tasks. The model learns from the relatively larger negative cases and can identify the negative cases more successfully.

The seven human subjects exhibited an accuracy of less than 75\% in the tasks of deep wound and infected wound. It can be observed that it is difficult for a human to perform these two tasks using only an image. The same trend is observed in the performance of the proposed model in the tasks of deep wound and infected wound. Among all of the medical personnel, resident B, who is the most junior among doctors, has the worst performance on average and performs a large performance gap compared to others on the infected wound and venous ulcer. It shows the difficulty of identifying those wound problems using only visual clues and the importance of long time-training for experienced wound experts. A comparison of the proposed model and human medical personnel using Cohen’s kappa difference indicates the promising performance of the proposed model. The proposed model performed better than three out of seven human medical personnel in the tasks of deep wound; four out of seven in the tasks of infected wound; two out of seven in the tasks of arterial and venous; one out of seven in the tasks of pressure wounds. In summary, when compared with human medical personnel, our model performed significantly better than or is non-inferior to all the medical personnel in all the five wound classification tasks.

Our model not only performs well in wound classification but does not require manual feature selection, as it is based on an end-to-end deep learning architecture~\cite{ref5_wang2015unified,ref6_shenoy2018deepwound,ref7_goyal2018dfunet,ref8_zahia2018tissue,ref22_nilsson2018classification,ref23_ootadeep}. A comparison of our proposed model with other deep learning-based methods is presented in Electronic Supplementary Material. Our model either clearly exhibited the best performance (with p-value < 0.05) compared with the other methods or is among the best methods. The proposed model size is very compact, which is advantageous in terms of the computation resource requirement.  

\begin{figure}[t!]
\centerline{
	\hspace{3mm}\includegraphics[width=1.0\columnwidth]{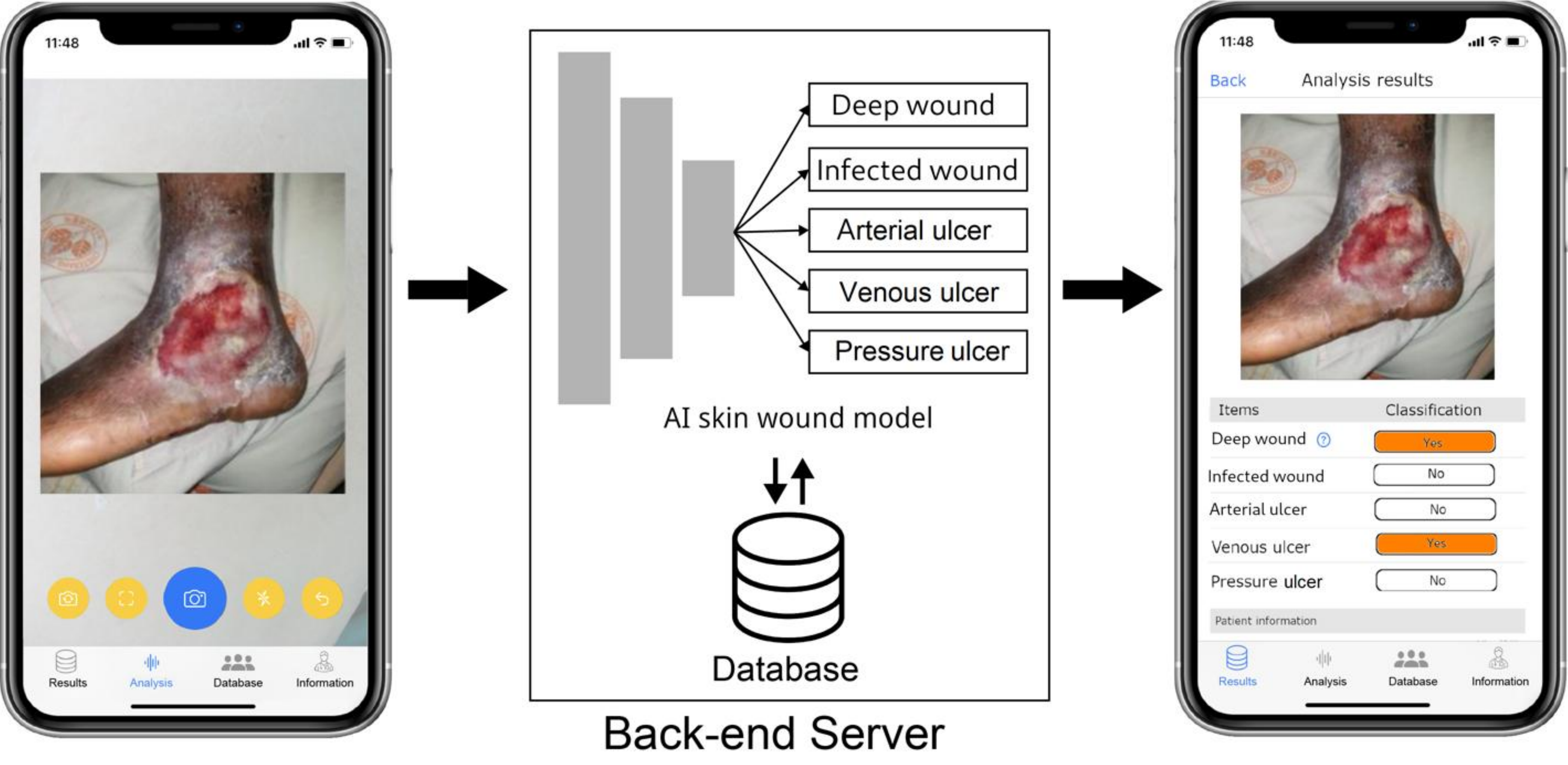}
}  
    \caption{Potential future clinical application. A user can take a photograph and our
wound assessment model will display the analysis results.}
    \vspace{-2mm}
	\label{fig:Figure5}
\end{figure}

In addition to the deep learning-based wound classification model, we plan to deploy such a core model to a web service prototype. The web service consists of a back-end computer server and a front-end smartphone app, as shown in Figure~\ref{fig:Figure5}. We plan to use such a prototype to study the applicability of the wound classification model in telemedicine. To perform wound assessment, a user can take a photograph of the wound using a smartphone. The wound image is then uploaded via a web service and fed to our deep learning-based wound classification model. The classification results are sent back to the mobile end and presented on the user interface of the mobile app. This mobile app can be extremely helpful and could be widely adopted by non-wound care medical personnel. In the future, we plan to extend our model to include more wound classification tasks to extend its scope. Patient information, such as vital signs and laboratory test results, will also be included to further improve our model performance.

The main limitation of our study is the generalizability of the proposed model. Although our various non-strict imaging conditions in capturing retrospective data already provide some generalizability for the proposed model in terms of coping with different imaging conditions, its prospective use may require some fine tuning with more data, particularly data captured under more diverse imaging conditions in different settings. Another limitation is the model uses only the wound images to classify the wound tasks which might omit some other non-visual information. We plan to incorporate additional information from the patient such as pain degree and temperature around the wound into the model to improve the model capacity.

In conclusion, we investigated a compact multi-task deep learning framework to classify deep wound, infected wound, arterial wound, venous wound, and pressure wound. The performance of the proposed model is good and is significantly better or non-inferior to all of the human study participants.

\section{Ethical approval}
The protocol has been approved by the Institutional Review
Board (1) of Taipei Veterans General Hospital. TPEVGH IRB
No.: 2020-08-008A.
\section{Conflict of Interest}
None
\section{Acknowledgments}
This study was approved by the Institutional Review Board
of Taipei Veterans General Hospital, approval number:
2020-08-008A.

This work was made possible by funding from Taipei
Veterans General Hospital and Invectec Corporation.
{\small
\bibliographystyle{ieee_fullname}
\bibliography{egbib}

\begin{thebibliography}{10}\itemsep=-1pt

\bibitem{ref4_anisuzzaman2022image}
DM Anisuzzaman, Chuanbo Wang, Behrouz Rostami, Sandeep Gopalakrishnan, Jeffrey
  Niezgoda, and Zeyun Yu.
\newblock Image-based artificial intelligence in wound assessment: A systematic
  review.
\newblock {\em Advances in Wound Care}, 11(12):687--709, 2022.

\bibitem{ref17_deng2009imagenet}
Jia Deng, Wei Dong, Richard Socher, Li-Jia Li, Kai Li, and Li Fei-Fei.
\newblock Imagenet: A large-scale hierarchical image database.
\newblock In {\em 2009 IEEE conference on computer vision and pattern
  recognition}, pages 248--255. Ieee, 2009.

\bibitem{ref19_efron1994introduction}
Bradley Efron and Robert~J Tibshirani.
\newblock {\em An introduction to the bootstrap}.
\newblock CRC press, 1994.

\bibitem{ref12_gao2019nddr}
Yuan Gao, Jiayi Ma, Mingbo Zhao, Wei Liu, and Alan~L Yuille.
\newblock Nddr-cnn: Layerwise feature fusing in multi-task cnns by neural
  discriminative dimensionality reduction.
\newblock In {\em Proceedings of the IEEE/CVF conference on computer vision and
  pattern recognition}, pages 3205--3214, 2019.

\bibitem{ref7_goyal2018dfunet}
Manu Goyal, Neil~D Reeves, Adrian~K Davison, Satyan Rajbhandari, Jennifer
  Spragg, and Moi~Hoon Yap.
\newblock Dfunet: Convolutional neural networks for diabetic foot ulcer
  classification.
\newblock {\em IEEE Transactions on Emerging Topics in Computational
  Intelligence}, 4(5):728--739, 2018.

\bibitem{ref3_han2017chronic}
George Han and Roger Ceilley.
\newblock Chronic wound healing: a review of current management and treatments.
\newblock {\em Advances in therapy}, 34:599--610, 2017.

\bibitem{ref16_he2016deep}
Kaiming He, Xiangyu Zhang, Shaoqing Ren, and Jian Sun.
\newblock Deep residual learning for image recognition.
\newblock In {\em Proceedings of the IEEE conference on computer vision and
  pattern recognition}, pages 770--778, 2016.

\bibitem{ref11_liu2019end}
Shikun Liu, Edward Johns, and Andrew~J Davison.
\newblock End-to-end multi-task learning with attention.
\newblock In {\em Proceedings of the IEEE/CVF conference on computer vision and
  pattern recognition}, pages 1871--1880, 2019.

\bibitem{ref20_mcclish1989analyzing}
Donna~Katzman McClish.
\newblock Analyzing a portion of the roc curve.
\newblock {\em Medical decision making}, 9(3):190--195, 1989.

\bibitem{ref18_mchugh2012interrater}
Mary~L McHugh.
\newblock Interrater reliability: the kappa statistic.
\newblock {\em Biochemia medica}, 22(3):276--282, 2012.

\bibitem{ref13_meyerson2017beyond}
Elliot Meyerson and Risto Miikkulainen.
\newblock Beyond shared hierarchies: Deep multitask learning through soft layer
  ordering.
\newblock {\em arXiv preprint arXiv:1711.00108}, 2017.

\bibitem{ref10_misra2016cross}
Ishan Misra, Abhinav Shrivastava, Abhinav Gupta, and Martial Hebert.
\newblock Cross-stitch networks for multi-task learning.
\newblock In {\em Proceedings of the IEEE conference on computer vision and
  pattern recognition}, pages 3994--4003, 2016.

\bibitem{ref22_nilsson2018classification}
C~Aguirre Nilsson and Medina Velic.
\newblock Classification of ulcer images using convolutional neural networks,
  2018.

\bibitem{ref23_ootadeep}
Subba~Reddy Oota, Vijay Rowtula, Shahid Mohammed, Jeffrey Galitz, Minghsun Liu,
  and Manish Gupta.
\newblock A deep multi-modal method for patient wound healing assessment.

\bibitem{ref21_provost1998case}
Foster~J Provost, Tom Fawcett, Ron Kohavi, et~al.
\newblock The case against accuracy estimation for comparing induction
  algorithms.
\newblock In {\em ICML}, volume~98, pages 445--453, 1998.

\bibitem{ref9_ruder2017overview}
Sebastian Ruder.
\newblock An overview of multi-task learning in deep neural networks.
\newblock {\em arXiv preprint arXiv:1706.05098}, 2017.

\bibitem{ref2_sen2021human}
Chandan~K Sen.
\newblock Human wound and its burden: updated 2020 compendium of estimates.
\newblock {\em advances in wound care}, 10(5):281--292, 2021.

\bibitem{ref14_sener2018multi}
Ozan Sener and Vladlen Koltun.
\newblock Multi-task learning as multi-objective optimization.
\newblock {\em Advances in neural information processing systems}, 31, 2018.

\bibitem{ref6_shenoy2018deepwound}
Varun~N Shenoy, Elizabeth Foster, Lauren Aalami, Bakar Majeed, and Oliver
  Aalami.
\newblock Deepwound: Automated postoperative wound assessment and surgical site
  surveillance through convolutional neural networks.
\newblock In {\em 2018 IEEE International Conference on Bioinformatics and
  Biomedicine (BIBM)}, pages 1017--1021. IEEE, 2018.

\bibitem{ref1_singer1999cutaneous}
Adam~J Singer and Richard~AF Clark.
\newblock Cutaneous wound healing.
\newblock {\em New England journal of medicine}, 341(10):738--746, 1999.

\bibitem{ref5_wang2015unified}
Changhan Wang, Xinchen Yan, Max Smith, Kanika Kochhar, Marcie Rubin, Stephen~M
  Warren, James Wrobel, and Honglak Lee.
\newblock A unified framework for automatic wound segmentation and analysis
  with deep convolutional neural networks.
\newblock In {\em 2015 37th annual international conference of the ieee
  engineering in medicine and biology society (EMBC)}, pages 2415--2418. IEEE,
  2015.

\bibitem{ref8_zahia2018tissue}
Sofia Zahia, Daniel Sierra-Sosa, Begonya Garcia-Zapirain, and Adel Elmaghraby.
\newblock Tissue classification and segmentation of pressure injuries using
  convolutional neural networks.
\newblock {\em Computer methods and programs in biomedicine}, 159:51--58, 2018.

\bibitem{ref15_zhang2021survey}
Yu Zhang and Qiang Yang.
\newblock A survey on multi-task learning.
\newblock {\em IEEE Transactions on Knowledge and Data Engineering},
  34(12):5586--5609, 2021.

\end{thebibliography}
}

\setcounter{section}{0}
\renewcommand\thesection{\Alph{section}}

\newpage
\section{Supplementary}
The accuracy of our model compared with the naive classifier (based on the probability 
distribution of the training data), Shenoy et al~\cite{ref6_shenoy2018deepwound}, Aguirre et al~\cite{ref22_nilsson2018classification}, and Liu et al~\cite{ref11_liu2019end} on the test dataset. * indicates a significant difference (p values \textless 0.05). Our model is compact while exhibiting significantly better or comparable performance to those other methods.

\begin{table}[h]
\caption{Difference in Cohen’s kappa with 95\% confidence intervals for performance comparison of our model and medical personnel. * indicates significant difference.}
\label{table:supplement_table}
\centering
\resizebox{\columnwidth}{!}{
\begin{tabular}{@{}cccccccc@{}}
\toprule
Model size ratio & Model & Metrics & Deep wound & Infected wound & Arterial wound & Venous wound & Pressure wound \\
\midrule
1 & Ours & Accuracy & 73.9 & 68.5 & 86.4 & 96.0 & 90.6 \\
\multirow{2}{*}{0.00} & \multirow{2}{*}{Naïve classifier} & Accuracy & 54.0* & 55.6* & 66.9* & 93.9 & 79.1* \\
& & p-value & 4.16e-9 & 1.01e-4 & 4.29e-13 & 0.09 & 2.58e-7 \\
\multirow{2}{*}{4.14} &\multirow{2}{*}{Shenoy et al.~\cite{ref6_shenoy2018deepwound}} & Accuracy & 70.0 & 68.1 & 83.8 & 90.4* & 87.3* \\ 
& & p-value & 0.07 & 0.83 & 0.06 & 1.86e-5 & 0.03\\
\multirow{2}{*}{23.27} & \multirow{2}{*}{Aguirre et al.~\cite{ref22_nilsson2018classification}} & Accuracy & 69.5 & 68.3 & 85.7 & 92.5* & 90.4 \\
& & p-value & 0.07 & 0.93 & 0.65 & 3.78e-3 & 0.87 \\ 
\multirow{2}{*}{0.96} & \multirow{2}{*}{Liu et al.~\cite{ref11_liu2019end}} & Accuracy & 75.1 & 67.1 & 83.3* & 95.5 & 92.3 \\ 
& & p-value & 0.56 & 0.56 & 0.02 & 0.56 & 0.09 \\
\bottomrule
\end{tabular}
}
\end{table}

\end{document}